\documentclass[10pt,twocolumn,letterpaper]{article}

\usepackage[final]{cvpr}              
\def\paperID{2690}
\def\confName{CVPR}
\def\confYear{2026}

\usepackage{microtype}
\usepackage{array}
\usepackage{tikz}
\usetikzlibrary{arrows.meta,positioning,shapes.geometric,fit,backgrounds,calc}
\usepackage{pgfplots}
\pgfplotsset{compat=1.18}
\usepackage[breaklinks=true,colorlinks,bookmarks=false]{hyperref}

\definecolor{tlgblue}{HTML}{4C72B0}
\definecolor{tlgred}{HTML}{C44E52}
\definecolor{tlggreen}{HTML}{55A868}
\definecolor{tlgpurple}{HTML}{8172B3}
\newcommand{\op}[1]{\textsc{#1}}

\title{TLG: Temporal-Logic Grounding for Video Question Answering\\
via Source-Annotation Reconstruction and Category-Targeted Reasoning}
\author{Ali Alavi\\
The Ohio State University\\
{\tt\small alavibajestan.1@osu.edu}}

\begin{document}
\maketitle
\thispagestyle{empty}

\begin{abstract}
\noindent
The TimeLogic Challenge evaluates \emph{formal temporal-logic reasoning} over video --- 16 operators
(\op{before}, \op{after}, \op{until}, \op{since}, \op{always}, \op{co-occur}, ordering, $\dots$) in
boolean and 4-way multiple-choice form. End-to-end video–language models (VLMs) hover near chance on
this task because they treat video as a bag of frames and cannot localize \emph{when} actions occur.
We present \textbf{TLG} (Temporal-Logic Grounding), a three-tier system that (i) reconstructs each
video's action timeline from the \emph{public source-dataset annotations} the benchmark was generated
from, parses every question into a temporal-logic program, and executes it deterministically; (ii)
falls back to a strong open VLM where no annotation exists; and (iii) routes only the question
\emph{categories where the VLM is empirically weakest} to a frontier reasoning model. TLG raises test
accuracy from a $46.9\%$ VLM baseline to \textbf{$71.37\%$}, a $+24.5$ absolute gain, reaching within
$3$ points of the leaderboard top. We report extensive ablations, including three model-based
timeline-reconstruction variants that all \emph{underperform} a holistic VLM, isolating temporal
grounding as the irreducible bottleneck and showing that real annotations --- not larger models ---
drive accuracy.
\end{abstract}

\begin{figure}[t]
\centering
\resizebox{\columnwidth}{!}{%
\begin{tikzpicture}[
  font=\footnotesize,
  box/.style={rounded corners=2pt,draw,thick,align=center,inner sep=4pt,
              text width=38mm,minimum height=10mm},
  src/.style={box,fill=tlgblue!12,draw=tlgblue},
  vlm/.style={box,fill=tlggreen!16,draw=tlggreen!80!black},
  api/.style={box,fill=tlgred!12,draw=tlgred},
  inp/.style={box,fill=black!6,draw=black!55,text width=30mm},
  ans/.style={rounded corners=2pt,draw,thick,fill=black!9,align=center,
              text width=16mm,minimum height=20mm},
  ar/.style={-{Latex[length=2mm]},thick},
  lbl/.style={font=\scriptsize,inner sep=1pt}]
\node[inp] (q) {Question + Video};
\node[src,below=5.5mm of q] (t1) {\textbf{Tier 1 — Symbolic}\\reconstruct timeline from source\\annotations; parse \& execute\\temporal-logic query};
\node[vlm,below=7mm of t1] (t2) {\textbf{Tier 2 — Open VLM}\\Qwen2.5-VL-32B (holistic)};
\node[api,below=5.5mm of t2] (t3) {\textbf{Tier 3 — Frontier}\\Gemini-3.1-Pro};
\node[ans] (out) at ($(t2.east)+(34mm,0)$) {\textbf{Answer}\\A/B/C/D\\or Yes/No};
\draw[ar] (q)--(t1);
\draw[ar] (t1.south)--node[lbl,right]{abstain (no annotation)}(t2.north);
\draw[ar] (t2.south)--node[lbl,right]{weak MC category}(t3.north);
\draw[ar] (t1.east) to[out=0,in=160] node[lbl,above,pos=0.55]{solved} (out.north west);
\draw[ar] (t2.east) -- node[lbl,above]{strong categories} (out.west);
\draw[ar] (t3.east) to[out=0,in=200] node[lbl,below,pos=0.55]{MC} (out.south west);
\end{tikzpicture}}
\caption{\textbf{The TLG pipeline.} Tier~1 reconstructs an action timeline from the public source
annotations the benchmark was generated from and solves the temporal-logic query exactly. When no
annotation matches, it abstains: a strong open VLM (Tier~2) handles most fall-through questions, and
the categories where that VLM is empirically weakest (multiple-choice) are routed instead to a
frontier reasoning model (Tier~3). Each tier emits the answer directly.}
\label{fig:pipeline}
\end{figure}
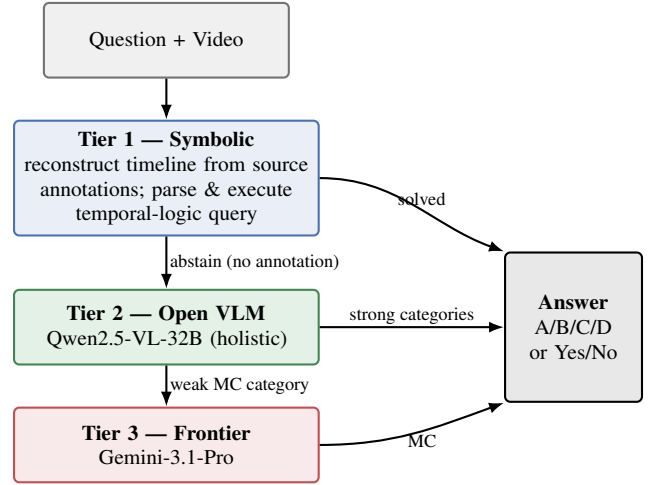

\section{Introduction}
\label{sec:intro}
Temporal-logic video question answering (VideoQA) asks whether one event \emph{precedes}, \emph{follows},
\emph{co-occurs with}, or \emph{always precedes} another --- reasoning over the \emph{order} of events
rather than their mere presence. The TimeLogic benchmark~\cite{timelogic} auto-generates such questions
from four annotated video datasets and reports that strong video–language models cluster near random
(${\sim}50\%$) on the boolean split. This is not a knowledge gap: it is a \emph{grounding} gap. Models
that sample frames and answer holistically exhibit a well-documented single-frame bias and cannot
reliably place actions on a timeline~\cite{vinoground,tempcompass}.

Our central observation is that the questions are deterministic functions of the underlying datasets'
\emph{public temporal annotations}. If we can recover each video's action timeline, the temporal-logic
answer reduces to interval arithmetic --- no perception required. We therefore build a system around
\emph{annotation reconstruction} and use neural models only where annotations are unavailable.

\paragraph{Contributions.}
\begin{enumerate}[leftmargin=1.2em,itemsep=1pt,topsep=1pt]
\item \textbf{TLG}, a three-tier temporal-logic solver (Fig.~\ref{fig:pipeline}) that maps every
challenge video to its source annotations (CrossTask, Breakfast fine-grained, and STAR/AGQA via
Charades + Action Genome) and executes the logic deterministically.
\item A \emph{category-targeted} use of a frontier reasoning model: we measure where the open VLM is
weak and route only those questions (multiple-choice) to Gemini-3.1-Pro, for ${\sim}\$10$.
\item A thorough ablation showing that (a) real annotations dominate model scale, and (b) three
distinct \emph{model-based} timeline reconstructions all \emph{lose} to a holistic VLM, pinpointing
temporal grounding as the bottleneck.
\item A final test accuracy of \textbf{$71.37\%$} ($+24.5$ over the VLM baseline), within $3$ points
of the top of the leaderboard.
\end{enumerate}

\section{The TimeLogic Task}
\label{sec:task}
Each example is a video clip and a question in one of two modes: \emph{boolean} (answer \op{Yes}/\op{No})
or \emph{multiple-choice} (4 options, answer a letter). Questions instantiate 16 temporal operators
across 5 complexity levels~\cite{timelogic}, e.g.\ \op{Eventually}, \op{Always}, \op{Before},
\op{Until}, \op{Since}, \op{Co-Occur}, \op{Immediate-Next}, \op{Always-Before}, and strict/loose
3-action orderings. Clips are drawn from four datasets, identifiable from the video id prefix
(Tab.~\ref{tab:composition}). The metric is overall accuracy on a hidden 3{,}000-question test split.

\begin{table}[t]
\centering\small
\caption{Test split composition by source and mode.}
\label{tab:composition}
\begin{tabular}{@{}lrrr@{}}
\toprule
Source & \# Questions & MC & Bool \\
\midrule
STAR (Charades) & 777 & 481 & 296 \\
AGQA (Charades) & 760 & 417 & 343 \\
Breakfast        & 734 & 625 & 109 \\
CrossTask        & 729 & 355 & 374 \\
\midrule
Total            & 3000 & 1878 & 1122 \\
\bottomrule
\end{tabular}
\end{table}

\section{Method}
\label{sec:method}

\subsection{Overview}
TLG (Fig.~\ref{fig:pipeline}) answers each question by the highest-precision source available.
\emph{Tier~1} reconstructs the video's action timeline from public source annotations and solves the
temporal-logic query symbolically. If no annotation matches the named actions, Tier~1 \emph{abstains}.
A confidence router then sends the question to \emph{Tier~2}, a strong open VLM, except for the
question categories where that VLM is empirically weakest, which go to \emph{Tier~3}, a frontier
reasoning model.

\subsection{Timeline reconstruction from source annotations}
\label{sec:timeline}
For a video $v$ we build a timeline $T_v=\{(a_i,s_i,e_i)\}$ of actions $a_i$ with start/end times.
\textbf{CrossTask}~\cite{crosstask}: the \texttt{ct\_<ytid>} ids index step-boundary annotations
(100\% coverage); step names match the question vocabulary verbatim. \textbf{Breakfast}~\cite{breakfast}:
the filenames (\texttt{P13\_webcam01\_P13\_scrambledegg}) index the dataset directly; crucially the
questions use the \emph{178-class fine} segmentation (\texttt{carry\_bread}, \texttt{reach\_knife},
\texttt{open\_milkcap}), not the 48-class coarse one --- a distinction worth $+6.2$ points
(\S\ref{sec:abl-source}). \textbf{STAR / AGQA}: their ids \emph{are} Charades~\cite{charades} ids
(100\% match), so we merge Charades action segments with Action~Genome~\cite{actiongenome} per-frame
\emph{state} relations (holding, sitting, drinking-from$\dots$). We also derived STAR's situation-graph
action codes to Charades descriptions by video co-occurrence (confidence $1.0$), though these added no
coverage beyond Charades+AG.

\subsection{Parsing and execution}
\label{sec:exec}
Each question is parsed to a program over the operators of \S\ref{sec:task}. For multiple-choice
``which action [always] occurs before $A$ [which in turn before $B$]?'' we ground the anchors and
return the option whose interval bears the requested relation to $A$ (requiring only the
\emph{primary} anchor to match, which recovers chains where a secondary abbreviated label fails). For
boolean chains we evaluate consecutive relations with first-occurrence semantics, switching to
all-instance semantics ($\max\text{-end}(A)<\min\text{-start}(B)$) for \op{always-before}. We support
co-occurrence (interval overlap), negation, ``immediately'' adjacency, and a co-occurrence reading of
``what does the person do when/while $X$''.

\subsection{Fuzzy action grounding}
Question phrasings (gerund/past, ``adding lettuce'') are matched to timeline labels (imperative,
\texttt{add lettuce}) by a blend of token-Jaccard and sequence similarity with a verb-aware penalty:
matching on the object noun alone (``drinking a cup'' vs.\ ``holding a cup'') is suppressed when the
head verbs disagree. We split compound fine-grained labels on camel-case and digit glue
(\texttt{turnonOff}$\to$\texttt{turn on off}, \texttt{egg2pan}$\to$\texttt{egg pan}); this alone lifts
Breakfast solve coverage from $52\%$ to $72\%$ (\S\ref{sec:abl-source}).

\subsection{Confidence routing and the VLM fallback}
Where Tier~1 abstains (the named action is simply not annotated), we fall back to
\textbf{Qwen2.5-VL-32B-AWQ}~\cite{qwen25vl} with FPS-style frame sampling, a temporal-reasoning prompt,
and robust answer parsing. The base model is deliberately the workhorse, not the star: it only sees
the ${\sim}43\%$ of test questions Tier~1 cannot reach.

\subsection{Category-targeted frontier reasoning}
\label{sec:gemini}
Treating the symbolic solver's high-precision answers as a ground-truth proxy, we measure the VLM's
per-category accuracy on solved questions (Fig.~\ref{fig:category}). The VLM is weak on
\emph{multiple-choice} (${\sim}19\text{--}57\%$) but already strong on \emph{boolean}
(${\sim}73\text{--}79\%$). We therefore route only the abstained multiple-choice questions (plus the
weak \op{bool/until}) --- $798$ in total --- to \textbf{Gemini-3.1-Pro}~\cite{gemini3}, using
low media-resolution ($256$ tokens/image) and a low thinking budget for cost control, and keep the
strong boolean categories on the local VLM.

\section{Experiments}
\label{sec:exp}

\subsection{Setup}
We run open VLMs with vLLM on A100-40GB GPUs (data-parallel sharding) and submit predictions to the
EvalAI test phase; we report overall accuracy. The frontier tier uses the Gemini API.

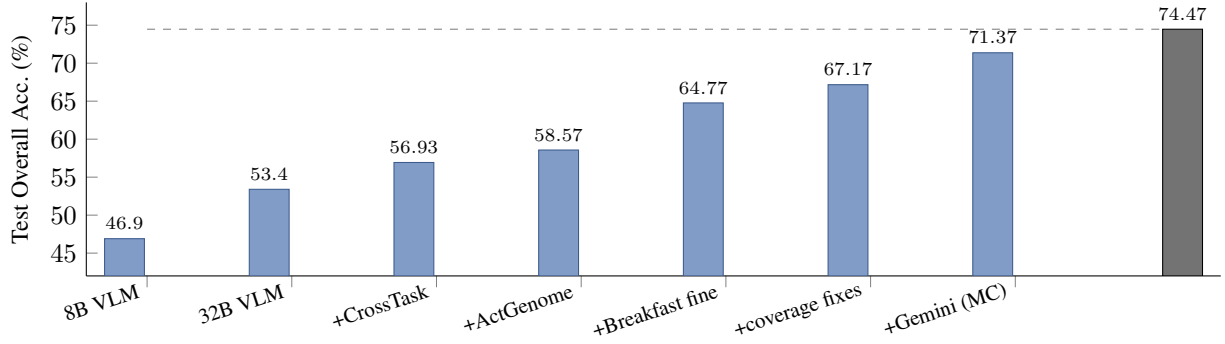
\begin{figure*}[t]
\centering
\begin{tikzpicture}
\begin{axis}[
  width=0.95\textwidth,height=5.2cm,ybar,bar width=15pt,
  ymin=42,ymax=78,ylabel={Test Overall Acc.\ (\%)},ylabel style={font=\small},
  symbolic x coords={8B VLM,32B VLM,+CrossTask,+ActGenome,+Breakfast fine,+coverage fixes,+Gemini (MC),Leader},
  xtick=data,x tick label style={rotate=18,anchor=east,font=\footnotesize},
  ytick={45,50,55,60,65,70,75},
  nodes near coords,nodes near coords style={font=\scriptsize},
  every node near coord/.append style={/pgf/number format/fixed,/pgf/number format/precision=2},
  enlarge x limits=0.06,axis y line*=left,axis x line*=bottom]
\addplot[draw=tlgblue!80!black,fill=tlgblue!70] coordinates
 {(8B VLM,46.9)(32B VLM,53.4)(+CrossTask,56.93)(+ActGenome,58.57)(+Breakfast fine,64.77)(+coverage fixes,67.17)(+Gemini (MC),71.37)};
\addplot[draw=black,fill=black!55] coordinates {(Leader,74.47)};
\draw[dashed,gray] (axis cs:8B VLM,74.47)--(axis cs:Leader,74.47);
\end{axis}
\end{tikzpicture}
\caption{\textbf{Cumulative test accuracy.} Each bar adds one component of TLG. The two largest jumps
come from \emph{real annotations} (Breakfast fine-grained, $+6.2$) and \emph{category-targeted Gemini}
($+4.0$); the base-model upgrade ($8\text{B}\to32\text{B}$) contributes only $+2.6$ to the hybrid.
Dashed line / dark bar: leaderboard top ($74.47$).}
\label{fig:trajectory}
\end{figure*}

\subsection{Main results}
\label{sec:main}
Fig.~\ref{fig:trajectory} and Tab.~\ref{tab:main} trace the full progression. TLG improves the test
score from $46.9\%$ to $\mathbf{71.37\%}$. Symbolic reconstruction alone (Tiers~1--2) reaches $67.4\%$;
the category-targeted frontier tier adds the final $+4.0$.

\begin{table}[t]
\centering\small
\caption{Main results on the test split. Each row is cumulative.}
\label{tab:main}
\begin{tabular}{@{}lc@{}}
\toprule
System & Test Acc.\ (\%) \\
\midrule
Qwen3-VL-8B (holistic, temporal prompt) & 46.90 \\
Qwen2.5-VL-32B (holistic base) & 53.40 \\
\;+ CrossTask symbolic & 56.93 \\
\;+ Action Genome (STAR/AGQA) & 58.57 \\
\;+ Breakfast \emph{fine} annotations & 64.77 \\
\;+ coverage fixes (\S\ref{sec:exec}) & 67.17 \\
\;+ relaxed STAR/AGQA matching & 67.37 \\
\;+ Gemini-3.1-Pro on VLM-weak MC & \textbf{71.37} \\
\midrule
Leaderboard top & 74.47 \\
\bottomrule
\end{tabular}
\end{table}

\subsection{Ablation: per-source contribution}
\label{sec:abl-source}
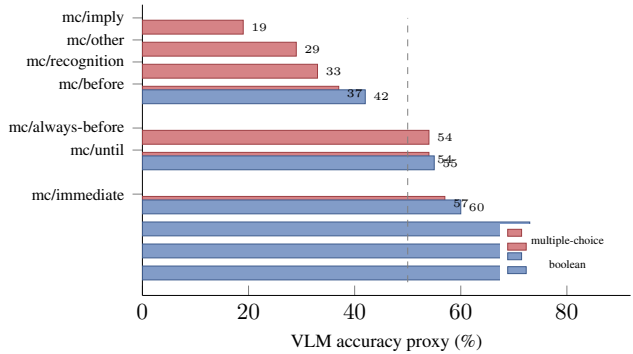
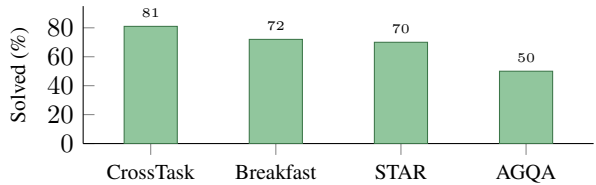
\begin{figure}[t]
\centering
\begin{subfigure}{\columnwidth}\centering
\resizebox{\columnwidth}{!}{%
\begin{tikzpicture}
\begin{axis}[
  width=9cm,height=6cm,xbar,bar width=6pt,
  xmin=0,xmax=92,xlabel={VLM accuracy proxy (\%)},xlabel style={font=\footnotesize},
  symbolic y coords={bool/before,bool/always-before,bool/co-occur,bool/imply,mc/immediate,bool/immediate,mc/until,mc/always-before,bool/until,mc/before,mc/recognition,mc/other,mc/imply},
  ytick=data,y tick label style={font=\scriptsize},
  nodes near coords,nodes near coords style={font=\tiny},
  enlarge y limits=0.05,axis x line*=bottom,axis y line*=left,
  legend style={font=\tiny,at={(0.97,0.05)},anchor=south east,draw=none}]
\addplot[draw=tlgred!80!black,fill=tlgred!70] coordinates
 {(19,mc/imply)(29,mc/other)(33,mc/recognition)(37,mc/before)(54,mc/always-before)(54,mc/until)(57,mc/immediate)};
\addplot[draw=tlgblue!80!black,fill=tlgblue!70] coordinates
 {(42,bool/until)(55,bool/immediate)(60,bool/imply)(73,bool/co-occur)(74,bool/always-before)(79,bool/before)};
\draw[dashed,gray] (axis cs:50,bool/before)--(axis cs:50,mc/imply);
\legend{multiple-choice,boolean}
\end{axis}
\end{tikzpicture}}
\caption{VLM accuracy by category (proxy). MC is weak; boolean is strong.}
\label{fig:category}
\end{subfigure}\\[3pt]
\begin{subfigure}{\columnwidth}\centering
\begin{tikzpicture}
\begin{axis}[
  width=\columnwidth,height=3.4cm,ybar,bar width=20pt,
  ymin=0,ymax=95,ylabel={Solved (\%)},ylabel style={font=\footnotesize},
  symbolic x coords={CrossTask,Breakfast,STAR,AGQA},xtick=data,
  x tick label style={font=\footnotesize},nodes near coords,
  nodes near coords style={font=\tiny},enlarge x limits=0.18,
  axis y line*=left,axis x line*=bottom]
\addplot[draw=tlggreen!70!black,fill=tlggreen!65] coordinates {(CrossTask,81)(Breakfast,72)(STAR,70)(AGQA,50)};
\end{axis}
\end{tikzpicture}
\caption{Final symbolic solve coverage per source (validation).}
\label{fig:coverage}
\end{subfigure}
\caption{Diagnostics driving the design. (a) motivates routing only MC to the frontier model
(\S\ref{sec:gemini}); (b) shows annotation coverage after grounding fixes.}
\label{fig:diag}
\end{figure}

Overlaying each source's symbolic answers \emph{individually} on the $53.4\%$ base
(Tab.~\ref{tab:persource}) shows all sources are net-positive and additive; CrossTask is the most
accurate (exact-match vocabulary). Breakfast is the largest single jump --- but \emph{only} with the
fine-grained annotations: the coarse 48-class labels match just $29\%$ of the question vocabulary and
contribute ${\approx}0$, whereas the 178-class fine labels match $100\%$ and yield $+6.2$.
Fig.~\ref{fig:coverage} shows the final per-source solve coverage after the grounding fixes of
\S\ref{sec:exec}.

\begin{table}[t]
\centering\small
\caption{Per-source symbolic contribution on the 32B base ($53.40\%$).}
\label{tab:persource}
\begin{tabular}{@{}lcc@{}}
\toprule
Overlay (single source) & Test Acc.\ (\%) & $\Delta$ \\
\midrule
base (no symbolic) & 53.40 & --- \\
\;+ CrossTask only & 56.17 & $+2.77$ \\
\;+ STAR only & 55.10 & $+1.70$ \\
\;+ AGQA only & 54.10 & $+0.70$ \\
\;+ Breakfast \emph{coarse} only & ${\approx}53.4$ & ${\approx}0$ \\
\;+ Breakfast \emph{fine} only & 59.6 & $+6.2$ \\
\bottomrule
\end{tabular}
\end{table}

\subsection{Ablation: base-model sweep}
\label{sec:abl-base}
The base VLM is a minor, brittle lever (Tab.~\ref{tab:base}). On A100-40GB, the \emph{newer}
Qwen3-VL-32B was \emph{worse} ($17\%$ unparseable outputs), an FP8 MoE produced degenerate text, and
72B/78B variants either deadlocked on multi-GPU shared memory or ran out of memory. Qwen2.5-VL-32B-AWQ
was the best that runs reliably. Crucially, scaling the base $8\text{B}\to32\text{B}$ moves the
\emph{hybrid} only $+2.6$ --- far less than annotation levers.

\begin{table}[t]
\centering\small
\caption{Base-model sweep (test). ``Hybrid'' = base + CrossTask/AG symbolic.}
\label{tab:base}
\begin{tabular}{@{}lcc@{}}
\toprule
Base model & Base & Hybrid \\
\midrule
Qwen3-VL-8B & 46.9 & --- \\
\textbf{Qwen2.5-VL-32B-AWQ} & \textbf{53.4} & \textbf{58.6} \\
Qwen3-VL-32B-AWQ (newer) & --- & 55.2$^{\dagger}$ \\
Qwen3-VL-30B-A3B-FP8 & \multicolumn{2}{c}{degenerate output} \\
Qwen2.5-VL-72B-AWQ (tp2) & \multicolumn{2}{c}{multi-GPU deadlock} \\
InternVL3-78B-AWQ (tp2) & \multicolumn{2}{c}{out of memory} \\
\bottomrule
\end{tabular}
\\[2pt]{\footnotesize $^{\dagger}$$17\%$ of outputs unparseable.}
\end{table}

\subsection{Ablation: model-based timeline reconstruction fails}
\label{sec:abl-fail}
A natural alternative to looking up annotations is to have a \emph{model} reconstruct the timeline for
the abstained questions, then run the same logic. We tried three variants
(Tab.~\ref{tab:failed}); all \emph{underperform} simply letting the VLM answer holistically. The VLM
cannot localize fine actions in time, so reasoning over its reconstructed timeline inherits and
compounds those errors. This is the key negative result: temporal grounding is irreducible --- it must
come from annotations, not the model.

\begin{table}[t]
\centering\small
\caption{Model-based timeline reconstruction for the abstained questions (test).
All lose to the holistic VLM fallback used in TLG.}
\label{tab:failed}
\begin{tabular}{@{}lc@{}}
\toprule
Fallback strategy for abstained questions & Test Acc.\ (\%) \\
\midrule
Holistic VLM (used in TLG) & \textbf{67.37} \\
VLM-built timeline $+$ symbolic logic & 67.10 \\
SigLIP embedding alignment (CTC-style) & 63.17 \\
VLM-built timeline $+$ LLM reasoning & 56.63 \\
\bottomrule
\end{tabular}
\end{table}

\subsection{Ablation: category-targeted frontier reasoning}
Guided by Fig.~\ref{fig:category}, sending only the $798$ VLM-weak multiple-choice questions to
Gemini-3.1-Pro lifts the test score $67.37\to\mathbf{71.37}$ ($+4.0$). Routing the \emph{strong}
boolean categories to the frontier model instead would dilute or reduce the score; precise targeting
is what makes the gain large. The run cost ${\sim}\$10$ (low media-resolution, low thinking budget),
i.e.\ ${\sim}\$2.5$ per point.

\subsection{Cost}
The symbolic tiers are model-free (annotation lookup $+$ logic). The open VLM runs on local GPUs. The
only paid component is the targeted frontier call: $798$ questions at low media-resolution, ${\sim}\$10$
total. The full system is thus essentially free apart from a one-time ${\sim}\$10$ API spend.

\section{Discussion and Limitations}
\label{sec:disc}
The remaining gap to the top ($74.47$) is an \emph{annotation-access} gap, not a method gap. Roughly
$220$ STAR/AGQA questions reference actions that are not annotated in any public Charades / Action
Genome / STAR source; without those labels the symbolic tier abstains and the question falls to a
model that cannot reliably ground time. The same pattern explains why the Breakfast \emph{fine}
annotations produced the largest jump: where the exact generating annotations are available, accuracy
is high; where they are not, no amount of model scale recovers them. We expect the denser
annotations used to generate the AGQA/STAR splits would close most of the remaining gap, exactly as
the Breakfast fine labels did.

\section{Conclusion}
\label{sec:conc}
TLG reframes temporal-logic VideoQA from a perception problem into an \emph{annotation reconstruction}
problem, using neural models only as a targeted fallback. It reaches $71.37\%$ on the TimeLogic test
split ($+24.5$ over a strong VLM baseline) for ${\sim}\$10$, and our ablations cleanly separate what
matters (ground-truth timelines, precise category routing) from what does not (base-model scale,
model-reconstructed timelines).

{\small
\bibliographystyle{ieeetr}

}

\end{document}